\documentclass{article}
\usepackage{tikz}
\usepackage{circuitikz}
\usetikzlibrary{arrows.meta}
\usetikzlibrary{fit}

\usepackage{microtype}
\usepackage{graphicx}
\usepackage{subcaption}
\usepackage{booktabs} 
\usepackage{multirow} 
\usepackage{adjustbox}
\usepackage{soul}
\usepackage{graphicx}
\usepackage{calc}   
\usepackage{caption}

\usepackage{hyperref}

\usepackage[accepted]{icml2026}

\usepackage{amsmath}
\usepackage{amssymb}
\usepackage{mathtools}
\usepackage{amsthm}

\usepackage[capitalize,noabbrev]{cleveref}

\theoremstyle{plain}

\theoremstyle{definition}

\theoremstyle{remark}

\usepackage[disable,textsize=tiny]{todonotes}

\icmltitlerunning{Does AI Assistance Preserve or Collapse Disagreement?}

\begin{document}

\twocolumn[
  \icmltitle{
  Does AI Assistance Preserve or Collapse Disagreement?\\A Study of Pre-Annotations in Ambiguous Video Labeling
}



  \icmlsetsymbol{equal}{*}

  \begin{icmlauthorlist}
    \icmlauthor{Juan Gutiérrez}{IPTC,UPM}
    \icmlauthor{Victor Gutiérrez-García}{UPM}
    \icmlauthor{José Luis Blanco-Murillo}{IPTC,UPM}
  \end{icmlauthorlist}

  \icmlaffiliation{IPTC}{Information Processing and Telecommunications Center}
  \icmlaffiliation{UPM}{ETSI de Telecomunicación, Universidad Politécnica de Madrid, Spain}

  \icmlcorrespondingauthor{}{juan.gutierrez@upm.es}

  \icmlkeywords{Human-in-the-Loop (HITL), Evaluation Methodology, Video Annotation, Cross-Modal Learning}

  \vskip 0.3in
]



\printAffiliationsAndNotice{}  

\begin{abstract}
AI-generated Pre-Annotations can accelerate video labeling, but they may also anchor annotators to model priors and suppress disagreement that is valuable for pluralistic dataset construction. We study this tradeoff in ambiguous temporal video annotation, where annotators choose event boundaries and assign context-dependent labels such as ``normal'' or ``abnormal.'' We introduce a controlled audit protocol that separates annotation cost, consensus alignment, inter-annotator consistency, temporal-boundary variation, semantic-label variation, latent-space standardization, and edit behavior.

In a counterbalanced pilot study with 18 annotators and 180 annotation sessions, a fixed CLIP-based Pre-Annotation engine reduced mean annotation time by 23.11\%; 72\% of annotators were faster with assistance, with a median per-annotator gain of 35\%. Assistance increased inter-annotator consistency and CLIP-space standardization, while semantic-label entropy changed only slightly in aggregate. A six-annotator human consensus diagnostic showed comparable alignment across conditions (AMI $\approx 0.64$ in both conditions), but this diagnostic is self-including and should be interpreted as descriptive rather than as independent evidence that model influence is absent. Overall, the results suggest that Pre-Annotations acted mainly as editable temporal scaffolds in this pilot setting, while leaving open the possibility of subtler semantic anchoring effects. We contribute an audit framework and anonymized interaction traces for studying when AI-assisted annotation preserves, reshapes, or collapses human disagreement.\end{abstract}

\section{Introduction}

Pluralistic alignment shifts the goal of AI development from optimizing toward a single aggregated preference to representing, preserving, and reasoning about the diversity of human judgments. This shift is especially important in data collection pipelines, where disagreement is often collapsed through majority voting, adjudication, or averaging \citep{uma2021learning, davani2022dealing}. In subjective or context-dependent tasks, disagreement can reflect different interpretations, experiences, assumptions, or values rather than simple annotation error \citep{fazelpour2025value}. Recent work argues that alignment datasets and benchmarks should preserve richer distributions of human judgments rather than assuming a unique ground truth \citep{gordon2022jury, sorensen2024roadmap}.

We study this issue in dense temporal video annotation. Unlike static image labeling, video annotation requires annotators to decide when events begin and end, how temporally extended activities should be grouped, and how observed behavior should be classified. Decisions are often ambiguous: event boundaries can be gradual, multiple actors may interact, and the meaning of an event depends on its context. In anomaly-oriented video labeling, the distinction between ``normal'' and ``abnormal'' behavior is especially sensitive to interpretation, social context, and the operational definition given to annotators \citep{Sultani_2018_CVPR,wu2023vadclip}.

AI-assisted annotation alters this decision process. Model-generated Pre-Annotations can reduce the mechanical burden of drawing segments from scratch, shifting the annotator's role from creation to verification and refinement. Yet this same assistance can also act as an anchor. If annotators accept proposed boundaries or labels too readily, the resulting dataset may appear cleaner while losing disagreement that would otherwise be informative for pluralistic modeling. The key methodological challenge is therefore not simply to measure whether assistance improves speed or agreement, but to determine what kind of agreement it increases.

We therefore ask how AI-generated Pre-Annotations change the distribution of human annotations in ambiguous temporal video labeling. Do they merely reduce mechanical effort and temporal jitter, or do they anchor annotators toward model priors and suppress meaningful disagreement? We answer this question with a controlled A/B study in which 18 annotators label high-ambiguity video clips with and without AI-generated Pre-Annotations. The Pre-Annotation engine is used as a fixed intervention that supplies temporally localized suggestions. We instrument the annotation interface to collect final labels, timing information, model suggestions, and interaction traces.

Our evaluation addresses three axes that are often conflated. First, we measure annotation cost through time-on-task. Second, we measure consensus alignment by comparing each annotation to a human consensus diagnostic, while avoiding the assumption that this reference is objective ground truth. Third, we measure disagreement and standardization through pairwise inter-annotator consistency and latent-space structure. This separation allows us to distinguish a reduction in arbitrary temporal boundary jitter from a collapse of meaningful annotator variation.

Our contributions are:
\begin{itemize}
    \item \textbf{A disagreement-centered framing of AI-assisted annotation}: we treat Pre-Annotations as behavioral interventions on the distribution of human annotations, rather than only as tools for improving throughput.
    \item \textbf{A controlled audit protocol} that separates annotation cost, consensus alignment, inter-annotator consistency, temporal-boundary variation, semantic-label variation, latent-space standardization, and edit-trace behavior.
    \item \textbf{A counterbalanced pilot study with 18 annotators and 180 annotation sessions} on ambiguous temporal video labeling.
    \item \textbf{Empirical evidence that assistance improves speed and increases standardization}: Pre-Annotations reduce mean annotation time by 23.11\% and increase pairwise consistency. A self-including human consensus diagnostic shows comparable aggregate alignment across conditions, which we report as descriptive evidence rather than as an independent test of whether model influence is absent.
    \item \textbf{A planned release of anonymized interaction traces and analysis code} to support follow-up work on pluralistic data collection, anchoring effects, and human--AI annotation workflows.
\end{itemize}

\section{Related Work}

\paragraph{Pluralistic alignment and annotation disagreement.}
Pluralistic alignment studies how AI systems can account for heterogeneous human values, preferences, and judgments rather than optimizing toward a single averaged target \citep{sorensen2024roadmap}. Annotation disagreement is central to this problem. In many pipelines, multiple labels are reduced to a single value through majority vote. However, prior work shows that disagreement can be systematic and informative in subjective tasks \citep{uma2021learning,davani2022dealing}. Methods such as jury learning preserve dissenting voices and enable downstream models to reason over distributions of human judgment \citep{gordon2022jury}. Our work studies an upstream stage of this pipeline: how does AI assistance during annotation reshape the disagreement that pluralistic modeling aims to preserve?

\paragraph{AI-assisted annotation and human verification.}
AI-assisted annotation is commonly motivated by efficiency. A model proposes labels, regions, or temporal segments, and humans correct them through a verification interface. Prior work has explored verification-centric labeling, interaction-efficient interfaces, and video-specific annotation tools \citep{papadopoulos2016we,papadopoulos2017extreme,kuznetsova2021efficient,deng2021eventanchor,feng2023video2action,shrestha2023feva}. Production tools such as CVAT and Label Studio have made model-assisted labeling a practical workflow for dataset construction \citep{Labelstudio,cvat_ai_corporation_2023_4009388}. These systems are usually evaluated through throughput or final-label quality. We evaluate an additional question: how suggestions alter (improve or degrade) the structure of annotator disagreement.

\paragraph{Ambiguous video labeling.}
Dense temporal video annotation is a useful setting for studying disagreement because annotators must make both temporal and semantic decisions. Open-world anomaly datasets such as UCF-Crime contain untrimmed surveillance-style footage with low resolution, unconstrained viewpoints, gradual event transitions, and ambiguous behavioral categories \citep{Sultani_2018_CVPR}. Recent video anomaly detection work has adapted vision--language models to these datasets \citep{wu2023vadclip}, but the human annotation process remains difficult to evaluate since no single frame-perfect ground truth captures all reasonable interpretations. We use this ambiguity as the target condition for our audit.

\paragraph{Vision--language models as annotation scaffolds.}
Vision--language models such as CLIP provide transferable visual and textual embeddings that can be adapted to video tasks with modest computation \citep{radford_learning_2021}. Efficient CLIP-style video learners use frame-level encoding, temporal pooling, and targeted fine-tuning to produce video-text representations \citep{rasheed2023fine,10323592}. Our Pre-Annotation engine follows this resource-conscious approach and combines a CLIP-style encoder with hierarchical spherical $k$-means \citep{JSSv050i10}. The engine supplies a stable source of suggestions for the human-subjects audit.

\section{Evaluation and Analysis}
\label{sec:evaluation}

Figure~\ref{fig:audit_framework} summarizes the audit pipeline. We collect annotations in Assisted and Unassisted conditions, preserve disaggregated interaction traces, and evaluate the resulting data along three axes: annotation cost, consensus alignment, and disagreement/standardization.

\begin{figure*}[t]
    \centering
    \begin{tikzpicture}[
        node distance=1.1cm and 1.2cm,
        every node/.style={font=\sffamily\small, align=center},
        box/.style={draw, rounded corners, minimum width=2.6cm, minimum height=0.8cm, fill=white},
        metric/.style={draw, rounded corners, minimum width=2.7cm, minimum height=0.75cm, fill=gray!10},
        arrow/.style={-Stealth, thick}
    ]
    \node[box] (videos) {Ambiguous\\videos};
    \node[box, right=of videos, yshift=0.8cm] (assisted) {Assisted\\Pre-Annotations visible};
    \node[box, right=of videos, yshift=-0.8cm] (unassisted) {Unassisted\\raw footage};
    \node[box, right=2.0cm of assisted, yshift=-0.8cm] (traces) {Disaggregated\\annotation traces};
    \node[metric, right=of traces, yshift=1.0cm] (time) {Annotation\\cost};
    \node[metric, right=of traces] (consensus) {Consensus\\alignment};
    \node[metric, right=of traces, yshift=-1.0cm] (disagree) {Disagreement \&\\standardization};

    \draw[arrow] (videos) -- (assisted);
    \draw[arrow] (videos) -- (unassisted);
    \draw[arrow] (assisted) -- (traces);
    \draw[arrow] (unassisted) -- (traces);
    \draw[arrow] (traces) -- (time);
    \draw[arrow] (traces) -- (consensus);
    \draw[arrow] (traces) -- (disagree);
    \end{tikzpicture}
    \caption{Pluralistic annotation audit framework. We evaluate whether Pre-Annotations reduce effort while preserving, reshaping, or collapsing human disagreement.}
    \label{fig:audit_framework}
\end{figure*}

\subsection{Materials and Model Sanity Check}
\label{sec:datasets}

The video clips used in this study were sampled from a held-out pool of open-world videos (UCF-Crime \citep{Sultani_2018_CVPR}) containing both normal and abnormal events. The Pre-Annotation engine was trained and validated on the video-language datasets described in Appendix~\ref{appendix:encoder_eval}.

The selected fixed encoder achieved 89.68\% Acc@1 and 99.64\% Acc@5 in video-text matching. This result is used as a sanity check to ensure that the Pre-Annotation source produces semantically plausible video-text representations. It does not, by itself, validate the temporal quality of the proposed segments.

\paragraph{Pre-Annotation intervention.}
The Assisted condition displayed model-generated temporal segments with proposed semantic labels and binary normal/abnormal flags. Annotators could accept, edit, relabel, split, merge, or remove these suggestions before submission. The same fixed Pre-Annotation engine was used for all Assisted sessions. The engine was not updated during the study and did not adapt to annotator behavior. This makes the model output a stable experimental intervention rather than an adaptive collaborator. Screenshots and implementation details of the annotation interface are provided in Appendix~\ref{appendix:interface}.

\subsection{Disaggregated Annotations and Interaction Traces}

We collected anonymized annotator-level traces for all 180 annotation sessions. The traces include final temporal segments, class labels, binary normal/abnormal flags, model Pre-Annotations, session-level timing, and edit histories. We preserve these disaggregated records because pluralistic analysis requires more than an aggregated final label. It requires visibility into where annotators converged, when they disagreed, and how model suggestions changed the annotation process.

\paragraph{Study design.}
The study used a within-subject, counterbalanced design. Each of the 18 annotators labeled ten videos: five in the Assisted condition, where model-generated Pre-Annotations were visible, and five in the Unassisted condition, where annotators labeled raw footage from scratch. Annotators did not label the same video twice. Across the study, this produced 180 annotation sessions over 30 videos, with six independent annotations per video. Condition order and video assignment were counterbalanced to reduce order and learning effects. Each annotation session is therefore nested within both an annotator and a video, and we treat annotator identity and video identity as repeated-measures factors in the analysis. Additional documentation of the annotation task, annotator recruitment, platform choices, and participant-protection considerations is provided in Appendix~\ref{appendix:CrowdWorkSheets}.

Table~\ref{tab:dataset_stats_detailed} summarizes the collected traces. The released artifact is intended to support follow-up analyses of efficiency, revision behavior, anchoring, and disagreement preservation in AI-assisted annotation workflows.

\begin{table}[t]
    \centering
    \caption{Summary of the disaggregated annotation traces collected during the controlled study.}
    \label{tab:dataset_stats_detailed}
    \begin{tabular}{lc}
        \toprule
        \textbf{Metric} & \textbf{Value} \\
        \midrule
        \multicolumn{2}{l}{\textit{Annotation and Segments}} \\
        Total Annotators & 18 \\
        Total Annotation Sessions & 180 \\
        Total Labeled Segments & 699 \\
        \midrule
        \multicolumn{2}{l}{\textit{Semantics}} \\
        Mean Classes per Video & $3.01 \pm 0.95$ \\
        Segments with Binary Event Flag & 670 \\
        Abnormal Event Segments & 323 \\
        Normal Event Segments & 347 \\
        \bottomrule
    \end{tabular}
\end{table}

\subsection{Metrics}

Our evaluation separates the annotation cost, aggregate consensus alignment, inter-annotator consistency, and the structure of disagreement. This separation is important because higher agreement can reflect either a desirable reduction of arbitrary boundary jitter or an undesirable collapse of meaningful annotator variation.

\paragraph{Annotation cost.}
We measure time-on-task in seconds from initial video load to final submission. We report session-level means and standard deviations, per-annotator percentage changes between conditions, and a mixed-effects estimate of the Assisted-condition effect on log time. Since annotation sessions are repeated within annotators and videos, we treat the timing results as repeated-measures data rather than as independent observations.

\paragraph{Consensus alignment.}
For each video, we construct a human Consensus Reference from the six independent annotations assigned to that video. We compare individual annotations to this reference using Adjusted Mutual Information (AMI), Normalized Mutual Information (NMI), and V-measure \citep{rosenberg2007vmeasure,vinh2010information}. These metrics compare induced temporal label partitions while accounting for chance structure in cluster assignments. The Consensus Reference is used only as an aggregate descriptive diagnostic, not as objective ground truth. Because the evaluated annotation is included in the six-annotator reference, this diagnostic is self-including and may inflate alignment values. We therefore do not interpret comparable consensus alignment as proof that anchoring is absent; instead, we interpret it jointly with pairwise consistency, boundary dispersion, semantic-label entropy, latent-space standardization, and edit traces.

\paragraph{Temporal partition construction.}
To compute AMI, NMI, V-measure, and entropy-based diagnostics, we convert each temporal annotation into a discrete label sequence over fixed temporal bins of width $\Delta=25$ seconds. Each bin receives the semantic class label, binary normal/abnormal flag, or background label corresponding to the segment with the largest temporal overlap with that bin. If no segment overlaps the bin, the bin is assigned the background label. If an annotator creates overlapping segments with different labels, the bin is assigned the label of the segment with the largest overlap; remaining ties are resolved by selecting the segment with the earliest start time. These bin-level sequences induce temporal label partitions that can be compared across annotators and against the Consensus Reference. Unless otherwise stated, agreement metrics are computed per video and summarized across videos. We did not perform a bin-width sensitivity analysis in this pilot, so agreement and entropy values should be interpreted relative to this fixed discretization choice.

\paragraph{Avoiding consensus circularity.}
Consensus-alignment metrics are computed against a six-annotator human Consensus Reference for each video. For every video $v$, the reference is constructed from all six independent annotations assigned to that video, aggregating their temporal label partitions into a single human consensus diagnostic. This construction intentionally describes alignment with the observed annotation distribution, but it also introduces circularity: when an individual annotation is evaluated, that annotation contributes to the reference against which it is compared. As a result, consensus-alignment values are likely optimistic and may obscure subtle differences between conditions. We therefore treat consensus alignment as a descriptive diagnostic only. The central interpretation of the study rests on the pattern across diagnostics: timing, pairwise consistency, boundary dispersion, semantic-label entropy, latent-space standardization, and edit behavior.

\paragraph{Inter-annotator consistency.}
We measure pairwise agreement among annotators within each condition. Higher pairwise consistency can indicate useful coordination around event boundaries, but it can also indicate disagreement collapse. We therefore interpret pairwise consistency jointly with consensus alignment, boundary-level variation, semantic-label variation, and interaction traces.

\paragraph{Boundary-level disagreement.}
To distinguish boundary standardization from semantic convergence, we separately measure temporal boundary variation. Because annotators may produce different numbers of segments, we first match segments within each video using maximum temporal overlap subject to label compatibility. For each matched event, we compute the dispersion of start and end times across annotators:
\begin{align}
\sigma_{\mathrm{start}}(v,e) &= \mathrm{StdDev}\{s_{a,v,e}\}_{a=1}^{A},\\
\sigma_{\mathrm{end}}(v,e) &= \mathrm{StdDev}\{t_{a,v,e}\}_{a=1}^{A},
\end{align}
where $s_{a,v,e}$ and $t_{a,v,e}$ are the start and end times assigned by annotator $a$ to event $e$ in video $v$. Lower boundary dispersion in the Assisted condition indicates temporal standardization, but not necessarily semantic agreement. Unmatched segments are reflected separately through segment-count variation.

\paragraph{Semantic-label disagreement.}
We measure variation in event labels and binary normal/abnormal flags separately from boundary placement. For each video and condition, we compute the entropy of the empirical label distribution:
\begin{equation}
H_v = - \sum_{c \in \mathcal{C}} p_v(c) \log p_v(c).
\label{eq}
\end{equation}
where $p_v(c)$ is the proportion of temporal bins in video $v$ assigned to label $c$ across annotators in that condition. This diagnostic helps distinguish reduced boundary jitter from reduced diversity of semantic interpretation.

\paragraph{Latent-space standardization.}
We compute the Silhouette Score of human-labeled segments using CLIP image embeddings and cosine distance \citep{shahapure2020cluster,radford_learning_2021}. Because the same encoder family contributes to the Pre-Annotation engine, this metric measures coherence under the model's representational geometry. We therefore treat it as a standardization diagnostic rather than as an independent measure of semantic correctness.

\paragraph{Anchoring diagnostics.}
For Assisted sessions, we additionally compute interaction-level diagnostics from edit traces: the fraction of Pre-Annotation segments accepted unchanged, boundary edited, split, merged, deleted, or relabeled; the mean boundary displacement from the model suggestion to the final human segment; and the label-flip rate between the model suggestion and the final annotation. These metrics directly test whether annotators used Pre-Annotations as editable scaffolds or accepted them with minimal revision.

\paragraph{Statistical analysis.}
The session results are not independent. We analyze the effects of conditions using repeated-measures estimates. For timing, we fit a mixed-effects model of the form
\begin{align}
\log(\mathrm{time}_{a,v})
&= \beta_0
  + \beta_1 \, \mathrm{Assisted}_{a,v} \nonumber\\
&\quad + \beta_2 \, \mathrm{Order}_{a,v}
  + \beta_3 \, \mathrm{Duration}_v \nonumber\\
&\quad + u_a + w_v + \epsilon_{a,v}.
\end{align}
where $u_a$ is a random intercept for annotator and $w_v$ is a random intercept for video. We report the Assisted-condition coefficient as the primary repeated-measures estimate of annotation-time reduction. For agreement, disagreement, and standardization metrics, we report video-level summaries. Session-level $t$-tests and Wilcoxon tests are included only as descriptive checks, not as the primary inferential evidence, because they do not fully account for repeated measurements within annotators and videos.

\subsection{Annotation Cost}

\begin{figure}[t]
    \centering
    \includegraphics[height=5.5cm]{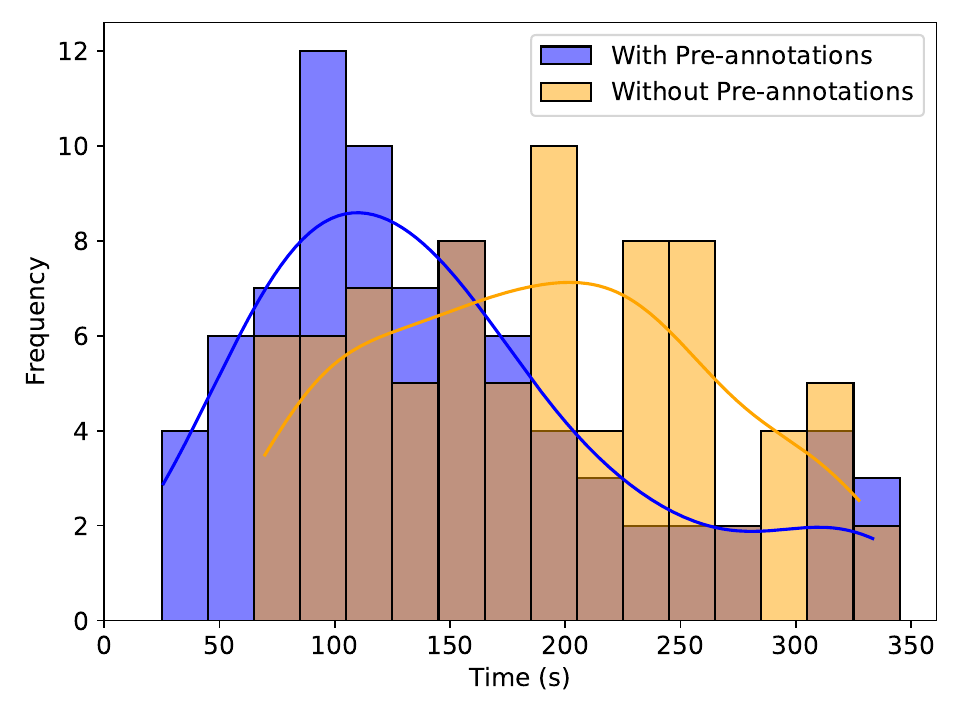}
    \caption{Distribution of annotation times with and without Pre-Annotations. Bin width is 20 seconds.}
    \label{fig:times_distributions}
\end{figure}

The first audit axis is annotation cost. We tracked the duration of the task, from initial video load to final submission, across all 180 sessions. Figure~\ref{fig:times_distributions} shows the time-on-task distributions for the Assisted (with-) and Unassisted (without-Pre-Annotations) conditions. Table~\ref{tab:time_results} reports the results on both parametric and non-parametric session-level statistical tests, which indicate a reliable timing difference in the collected sessions.

\begin{table}[t]
\centering
\caption{Annotation-time results. Time is measured in seconds from initial video load to final submission. Uncertainty is reported as approximate 95\% confidence intervals. The mixed-effects estimate is reported on log time and transformed into an implied percentage reduction.}
\label{tab:time_results}
\resizebox{\linewidth}{!}{%
\begin{tabular}{lccc}
\toprule
\textbf{Condition} & \textbf{Mean} & \textbf{Std.} & \textbf{95\% CI} \\
\midrule
Assisted & 146.42 & 79.28 & $\pm 16.60$ \\
Unassisted & 190.44 & 72.63 & $\pm 15.22$ \\
\midrule
\multicolumn{4}{l}{\textit{Effect estimates}} \\
Mean reduction & \multicolumn{3}{c}{23.11\%} \\
Mean difference & \multicolumn{3}{c}{$-44.02 \pm 22.44$ s} \\
Annotators faster with assistance & \multicolumn{3}{c}{72\%} \\
Median per-annotator gain & \multicolumn{3}{c}{35\%} \\
\midrule
\multicolumn{4}{l}{\textit{Repeated-measures estimate}} \\
Mixed-effects coefficient, $\hat{\beta}_1$ & \multicolumn{3}{c}{$-0.263\;[-0.397,\,-0.129]$} \\
Implied time reduction & \multicolumn{3}{c}{23.1\% $[12.1\%,\,32.8\%]$} \\
Wald test & \multicolumn{3}{c}{$z=-3.86,\ p<0.001$} \\
\midrule
\multicolumn{4}{l}{\textit{Session-level descriptive tests}} \\
$t$-test & \multicolumn{3}{c}{$t=-3.64,\ p=0.0004$} \\
Wilcoxon & \multicolumn{3}{c}{$W=853.0,\ p=0.0002$} \\
\bottomrule
\end{tabular}%
}
\end{table}

As shown in Table~\ref{tab:time_results}, the Assisted condition reduced the mean annotation time from 190.44 seconds to 146.42 seconds, corresponding to a 23.11\% reduction in mean time-on-task. Normalizing within annotator, 72\% of participants were faster with Pre-Annotations, with a median per-annotator gain of 35\%. A mixed-effects model with annotator and video random intercepts produced an Assisted-condition coefficient of approximately $\hat{\beta}_1=-0.263$ on log time, corresponding to an estimated 23.1\% reduction in annotation time. The session-level $t$-test and Wilcoxon test are reported only as descriptive checks because they do not fully account for repeated measurements within annotators and videos.

\begin{figure}[tb]
    \centering
    \begin{subfigure}{0.7\linewidth}
        \includegraphics[width=\linewidth]{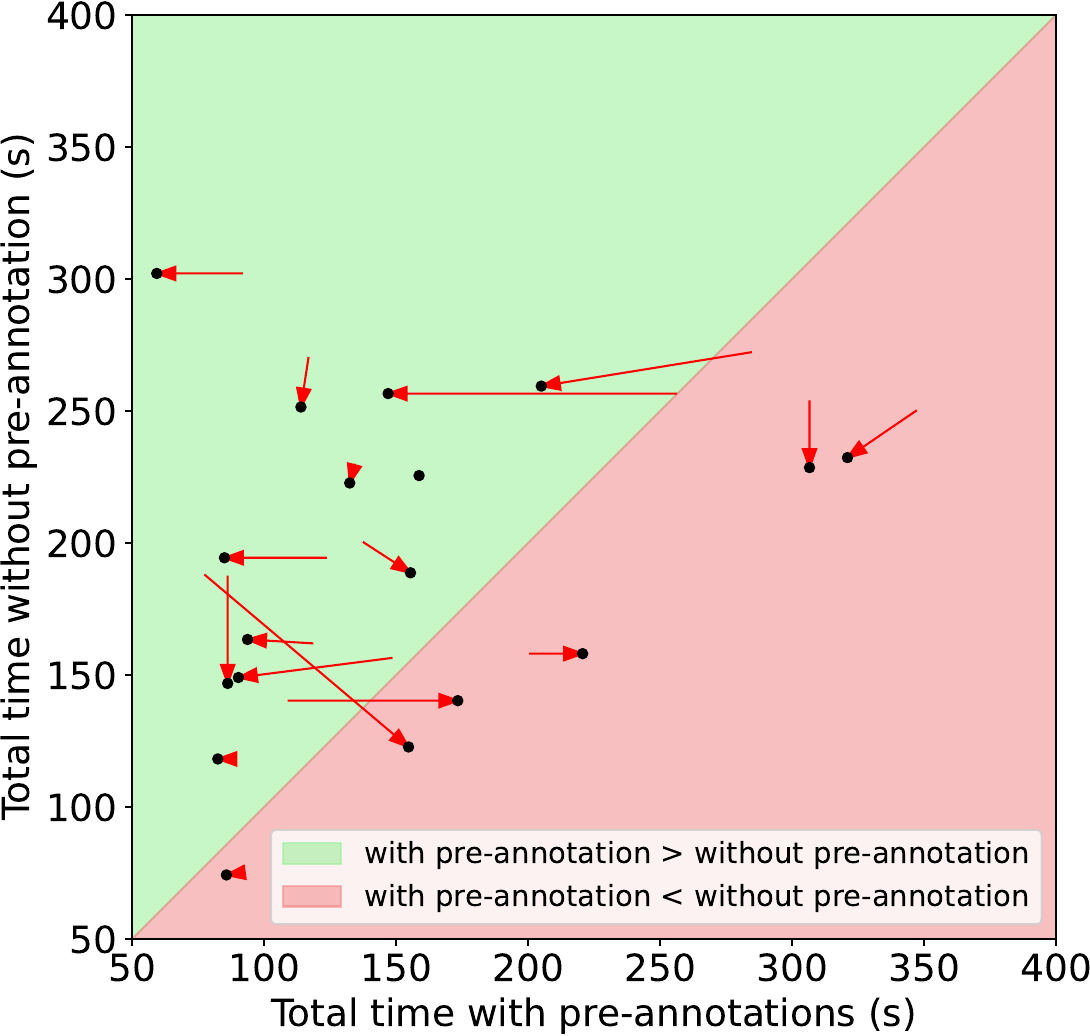}
        \subcaption{Normalized annotation time}
    \end{subfigure}
    \hfill
    \begin{subfigure}{0.7\linewidth}
        \includegraphics[width=\linewidth]{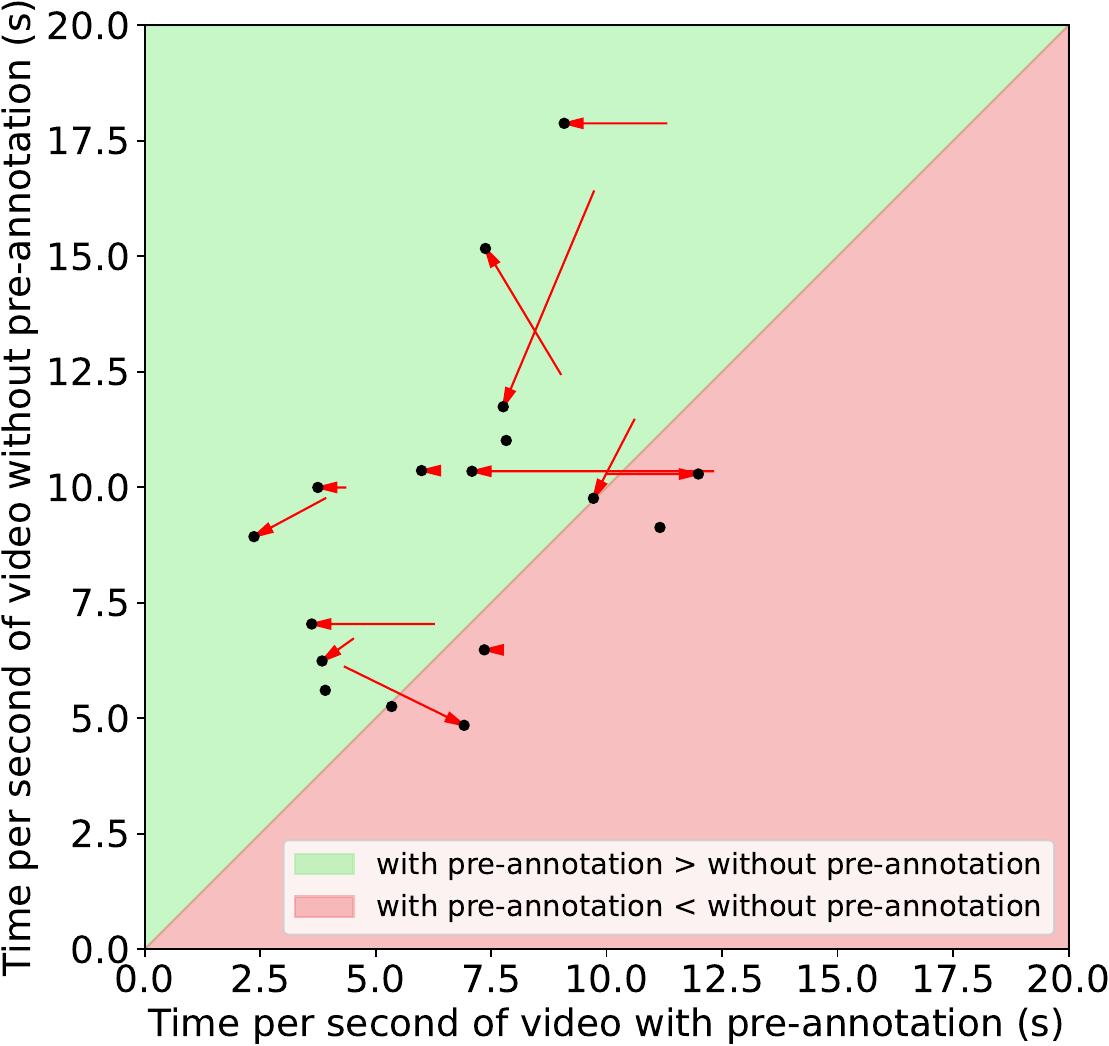}
        \subcaption{Normalized shift after training videos}
    \end{subfigure}
    \caption{Per-annotator efficiency effects before and after removing early familiarization trials. Each red arrow represents the change for an annotator after excluding the first four training videos. Times are normalized by video duration. Points in the assisted-faster region indicate annotators who completed annotations faster with Pre-Annotations.}
    \label{fig:efficiency_per_annotator}
\end{figure}

The per-annotator comparison in Figure~\ref{fig:efficiency_per_annotator} shows that the efficiency benefit is not driven solely by a small number of sessions. On these axes, the $x$-coordinate is Assisted time and the $y$-coordinate is Unassisted time; therefore, points above the diagonal indicate faster completion with Pre-Annotations. For each annotator, we removed the first four videos they annotated in chronological order and recomputed the per-annotator normalized timing comparison. The normalized comparison after the initial familiarization videos shows a similar pattern, suggesting that annotators became more efficient at reviewing and editing suggestions once they were familiar with the interface.

Annotator feedback suggests that Pre-Annotations mainly reduced the time spent deciding on an initial segmentation structure. In some cases, annotators reported that the suggested segments were sufficiently close to require only review and minor correction. In other cases, the suggestions required substantial editing. This variation matters for pluralistic annotation: the aim is not to make humans accept outputs quickly, but to reduce mechanical effort while preserving the opportunity for correction and disagreement.

\subsection{Consensus Alignment and Inter-Annotator Consistency}

The second audit axis asks whether AI assistance changes semantic judgment. In pluralistic annotation, higher agreement is ambiguous: it can indicate useful reduction of boundary jitter, but it can also indicate homogenization around model priors. We therefore separate two measurements: alignment with a six-vote human Consensus Reference and pairwise consistency among annotators within each condition.

Table~\ref{tab:disagreement_results} shows that Assisted and Unassisted annotations have nearly identical alignment with the Consensus Reference. The Assisted condition obtains AMI 0.64, while the Unassisted condition obtains AMI 0.63. NMI and V-measure are also comparable across conditions. However, this consensus diagnostic is self-including: each evaluated annotation contributes to the six-annotator reference against which it is compared. These values should therefore be interpreted as descriptive alignment with the observed annotation distribution, not as an independent validation that assistance preserves human judgment or eliminates anchoring.

Pairwise inter-annotator consistency is higher in the Assisted condition. Assisted annotators reached AMI 0.67, compared to AMI 0.62 for the Unassisted condition. This indicates standardization: annotators who see the same suggestions converge more strongly with each other. However, higher consistency is not, by itself, evidence that meaningful disagreement has been preserved. It may reflect a useful reduction of arbitrary temporal jitter, but it may also reflect anchoring toward model-proposed boundaries or labels. We therefore treat the increase in pairwise agreement as a signal requiring decomposition into boundary-level and semantic-label components, rather than as a standalone success criterion.

\begin{table}[t]
\centering
\caption{Consensus alignment and disagreement diagnostics. Positive deltas indicate larger assisted-condition values.}
\label{tab:disagreement_results}
\resizebox{\linewidth}{!}{%
\begin{tabular}{lccc}
\toprule
\textbf{Diagnostic} & \textbf{Assisted} & \textbf{Unassisted} & $\Delta$ \\
\midrule
Consensus AMI & 0.64 & 0.63 & +0.01 \\
Consensus NMI & 0.64 & 0.64 & 0.00 \\
Consensus V-measure & 0.64 & 0.64 & 0.00 \\
\midrule
Pairwise AMI & 0.67 & 0.62 & +0.05 \\
Pairwise NMI & 0.67 & 0.62 & +0.05 \\
Pairwise V-measure & 0.67 & 0.62 & +0.05 \\
\bottomrule
\end{tabular}%
}
\end{table}

\subsection{Boundary and Semantic Disagreement Decomposition}

To better interpret the increase in pairwise consistency, we decompose disagreement into temporal boundary variation and semantic-label variation. Boundary variation captures differences in the placement of annotators' event start and end times. Semantic-label variation captures differences in the class labels and normal/abnormal flags assigned to the segments.

This decomposition is important because the pluralistic impact of a disagreement depends on its source. Reduced boundary jitter may be desirable when annotators identify the same event but differ by a few frames or seconds. Reduced semantic-label entropy, by contrast, may indicate that model suggestions suppress alternative interpretations of ambiguous events.

Table~\ref{tab:disagreement_decomposition} reports three additional diagnostics: boundary dispersion, segment-count variation, and label entropy. Boundary dispersion measures the standard deviation of matched start and end times across annotators. Segment-count variation measures whether assistance causes annotators to produce a more uniform number of segments. Label entropy measures whether assistance reduces diversity in semantic labels or binary event flags.

\begin{table}[t]
\centering
\caption{Disagreement decomposition. Boundary-level metrics distinguish temporal standardization from semantic-label convergence. Values are computed per video and summarized across videos.}
\label{tab:disagreement_decomposition}
\resizebox{\linewidth}{!}{%
\begin{tabular}{lccc}
\toprule
Diagnostic & Assisted & Unassisted & Difference \\
\midrule
Start-boundary dispersion & 0.80 & 1.10 & -0.30 \\
End-boundary dispersion & 1.21 & 2.10 & -0.89 \\
Segment-count variance & 1.26 & 0.72 & +0.54 \\
Class-label entropy & 1.17 & 1.18 & -0.01 \\
Normal/abnormal entropy & 0.60 & 0.64 & -0.04 \\
\bottomrule
\end{tabular}
}
\end{table}

Interestingly, assistance reduced boundary dispersion for matched events but increased segment-count variance. This suggests that Pre-Annotations may standardize local boundary placement without fully standardizing higher-level segmentation granularity. Annotators may agree more closely on the timing of events they identify, while still differing in whether they split or merge adjacent activities. This reinforces the need to analyze boundary placement, event count, and semantic labels separately.

If the Assisted condition primarily reduces boundary dispersion while preserving comparable semantic-label entropy, this would support the interpretation that Pre-Annotations act as temporal scaffolds. If the Assisted condition also substantially reduces semantic-label entropy, this would provide stronger evidence of disagreement collapse or anchoring.

\subsection{Anchoring and Edit-Trace Diagnostics}

The central risk of Pre-Annotations is that they may anchor annotators to model priors. Consensus alignment and pairwise agreement provide indirect evidence of this risk, whereas interaction traces enable a more direct measurement. In the Assisted condition, we therefore analyze how annotators modified model-generated suggestions before submission.

We define a Pre-Annotation segment as \emph{accepted unchanged} if its temporal boundaries and label are submitted without modification. A segment is \emph{boundary edited} if its start or end time changes, \emph{relabeled} if its semantic class or binary event flag changes, \emph{split} or \emph{merged} if its temporal structure is changed, and \emph{deleted} if it is removed from the final annotation.

These edit categories are not mutually exclusive: a segment may be both boundary-edited and relabeled, for example. We therefore interpret the edit-trace diagnostics as evidence about revision behavior rather than as a partition of all possible edit operations.

\begin{table}[t]
\centering
\caption{Anchoring diagnostics from Assisted-session edit traces. These metrics distinguish passive acceptance of model suggestions from active human revision.}
\label{tab:anchoring_diagnostics}
\begin{tabular}{lc}
\toprule
Edit-trace diagnostic & Value \\
\midrule
Pre-Annotation segments accepted unchanged & 24.20\% \\
Boundary edited & 64.69\% \\
Relabeled & 50.12\% \\
Split or merged & 56.79\% \\
Deleted & 0.00\%\\
Mean start-boundary displacement & 1.36 s \\
Mean end-boundary displacement & 1.09 s \\
Normal/abnormal label-flip rate & 31.68\% \\
\bottomrule
\end{tabular}
\end{table}

Table~\ref{tab:anchoring_diagnostics} shows that only 24.20\% of Pre-Annotation segments were accepted unchanged, while 64.69\% were boundary edited, 50.12\% were relabeled, and 56.79\% were split or merged. The zero deletion rate indicates that annotators rarely removed model suggestions outright; instead, they primarily revised suggestions through boundary edits, relabeling, splitting, or merging. These diagnostics are necessary because higher agreement can arise through different mechanisms. If most Pre-Annotations are accepted unchanged, then the Assisted condition may reflect anchoring to model priors. If annotators frequently edit boundaries, relabel segments, split or merge suggestions, and delete incorrect proposals, then Pre-Annotations are better understood as editable scaffolds that reduce mechanical effort while preserving human oversight. Active editing does not rule out anchoring, since annotators may still make edits around model-proposed structures rather than constructing independent alternatives from scratch. The edit traces therefore rule out only a purely passive-acceptance account. They do not rule out subtler forms of anchoring, such as local adjustment around model-proposed temporal structures or selective retention of model-proposed semantic categories.

\subsection{Latent-Space Standardization}

The third audit axis measures whether annotations become more structurally coherent in the visual embedding space. We compute the Silhouette Score of human-labeled segments using CLIP image embeddings and cosine distance. Because the same encoder family is used in the Pre-Annotation engine, this metric is aligned with the model's representational geometry. We therefore interpret it as a standardization diagnostic: it measures whether assisted annotators produce boundaries that are more coherent under the embedding space used to generate suggestions.

\begin{figure}[t]
    \centering
    \includegraphics[width=0.7\linewidth]{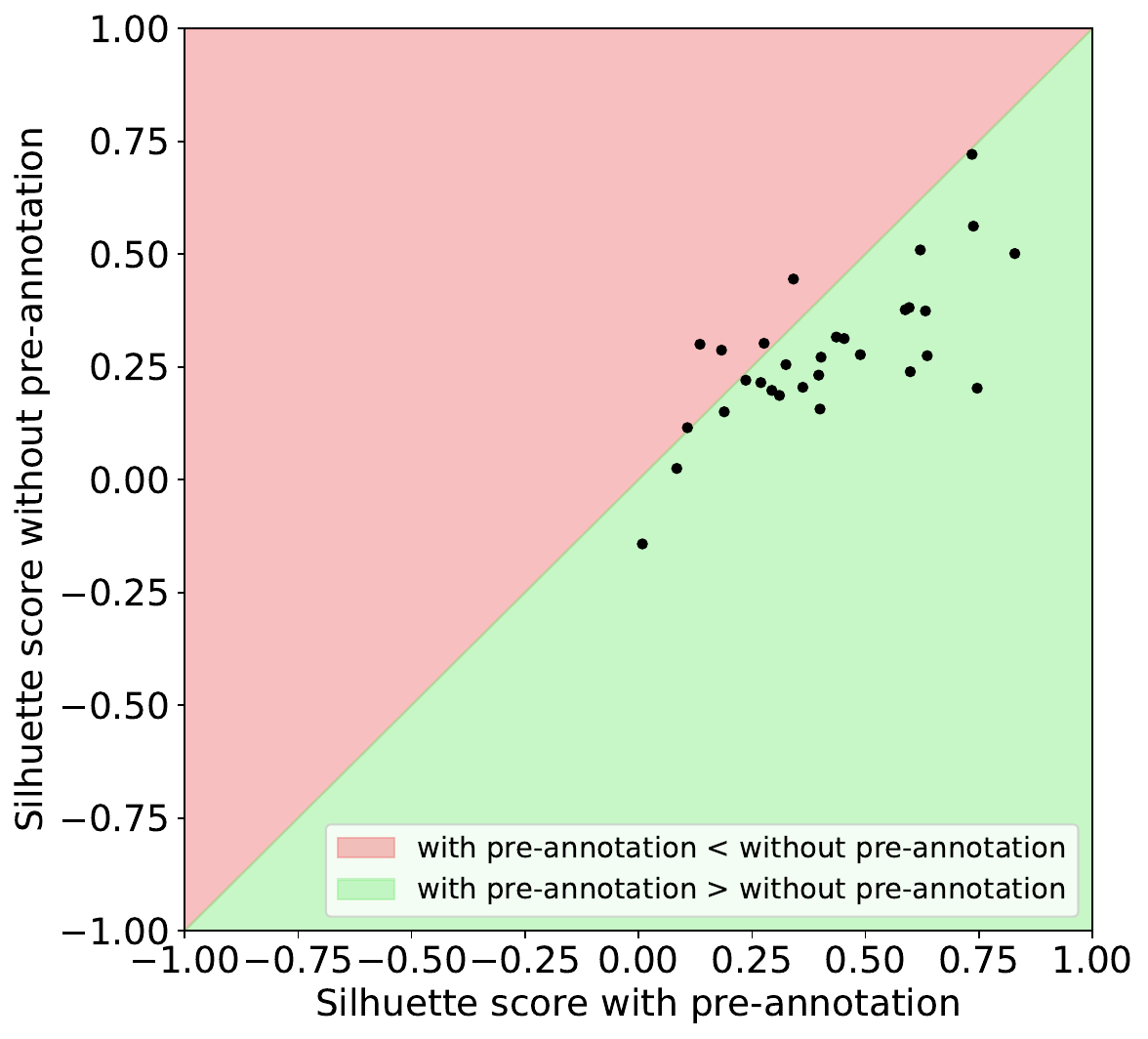}
    \caption{Latent-space standardization. Points below the diagonal indicate videos where Assisted annotations have higher Silhouette Scores than Unassisted annotations.}
    \label{fig:ss_scatter}
\end{figure}

Figure~\ref{fig:ss_scatter} shows that the Assisted condition yields higher Silhouette Scores in 25 of 30 videos. The mean score increases from 0.28 in the Unassisted condition to 0.41 in the Assisted condition, corresponding to an approximately 46.4\% relative increase when computed from the rounded means. This pattern indicates that Assisted annotations are more coherent under the representational geometry used by the Pre-Annotation engine.

However, this result should be interpreted conservatively. It is a diagnostic of model-space standardization. Higher latent-space coherence is desirable when it reflects reduced arbitrary boundary jitter, but problematic when it reflects convergence toward model priors. For this reason, we interpret the Silhouette result jointly with consensus alignment, boundary-level disagreement, semantic-label entropy, and edit-trace diagnostics rather than as standalone evidence of improved annotation quality.

\section{Discussion}

The results reveal a distinction that is central to pluralistic data collection. Pre-Annotations increased efficiency, pairwise inter-annotator consistency, and CLIP-space standardization. Semantic-label entropy changed only slightly in aggregate, and a six-annotator human consensus diagnostic produced comparable alignment values across conditions. However, the consensus diagnostic is self-including and therefore cannot establish that assistance preserves human judgment or that anchoring is absent. The strongest supported interpretation is more modest: in this pilot setting, assistance made annotation faster and more standardized, with no large observed aggregate collapse in semantic-label entropy under the reported fixed-bin diagnostics. Higher consistency is not automatically desirable: it may indicate useful reduction of arbitrary temporal boundary jitter, or it may indicate harmful collapse of meaningful disagreement. We therefore interpret the study as an audit of how AI assistance changes the structure of annotation, not as definitive evidence that all meaningful disagreement is preserved.

The strongest empirical summary is therefore not that Pre-Annotations preserve disagreement, but that they make annotation faster and more standardized without producing a large observed change in aggregate semantic-label entropy. The remaining question is what kind of standardization assistance induces. In our results, the largest changes appear in time-on-task, pairwise agreement, boundary dispersion, and CLIP-space Silhouette Score. In contrast, semantic-label entropy changes little, and consensus alignment appears comparable under a descriptive self-including diagnostic. This pattern is more consistent with temporal scaffolding than with wholesale semantic collapse, although the evidence is not sufficient to conclude that all meaningful disagreement was preserved.

A key ambiguity is that increased agreement can be interpreted either as boundary regularization or as anchoring. Our protocol addresses this ambiguity by separating annotation time, consensus alignment, pairwise consistency, latent-space standardization, boundary variation, semantic-label entropy, and edit behavior. The central empirical signal is therefore not a single metric, but a pattern across diagnostics. In the present pilot, the strongest evidence supports efficiency gains and standardization. Stronger claims about disagreement preservation require direct evidence that semantic-label diversity, minority interpretations, and reasonable alternative segmentations remain available after assistance.

This pattern supports a practical design principle: AI assistance should be evaluated as an intervention on the distribution of human labels, not only as a way to reduce cost. A workflow can appear successful under standard metrics because it is faster and cleaner, while still being problematic if it collapses disagreement that should remain visible. Conversely, some forms of standardization may be desirable when they reduce arbitrary boundary jitter and leave annotators free to correct semantic labels.

The interaction traces are central to this distinction. Final aggregate labels do not reveal whether annotators accepted model suggestions, edited them, rejected them, or converged independently. By preserving disaggregated traces, the audit makes it possible to analyze both the final label distribution and the process by which it was produced.

\section{Limitations}

This study has several limitations. First, the sample size is modest: 18 annotators, 30 videos, and six annotations per video. This design is suitable for a controlled pilot audit of one annotation workflow, but it is underpowered to rule out subtle shifts in semantic interpretation, minority-label suppression, or community-specific disagreement. A near-zero aggregate change in semantic-label entropy should therefore not be interpreted as proof that all meaningful disagreement was preserved.

Second, annotators were university-affiliated volunteers, so the results do not provide population-level evidence about broader pluralistic values. Judgments about normality, abnormality, event boundaries, and contextual relevance may vary across communities and stakeholder groups. The study evaluates how one AI-assisted workflow changes annotation behavior under controlled conditions; it does not establish how affected communities would interpret the same videos or labels.

Third, we evaluate one fixed Pre-Annotation engine and one interface. Results may differ with other models, suggestion qualities, interface designs, video domains, or annotator populations. Fourth, the Consensus Reference is an aggregate diagnostic rather than objective ground truth. Because the six-annotator reference includes the evaluated annotation, consensus-alignment values are self-including and may be optimistic. Comparable alignment therefore does not establish that anchoring is absent. Fifth, agreement and entropy metrics depend on the fixed temporal binning scheme used to discretize continuous annotations; we did not perform a bin-width sensitivity analysis in this pilot.

Finally, the task involves sensitive open-world video content. Larger studies should use paid participation, stratified recruitment, content warnings, opt-out and skip mechanisms, debriefing resources, and stakeholder-specific analysis. These results should not be used to justify deployment of surveillance or anomaly-detection systems without broader ethical review and stakeholder engagement.

\section{Conclusion}

We studied AI-assisted temporal video annotation as a pluralistic data-collection problem: model suggestions should be evaluated not only by whether they speed up labeling, but by whether they preserve, reshape, or collapse disagreement in ambiguous tasks.

In a controlled pilot with 18 annotators and 180 sessions, Pre-Annotations reduced annotation time for most participants and increased inter-annotator consistency. Boundary dispersion decreased, while semantic-label entropy changed only slightly in aggregate. A self-including human consensus diagnostic showed comparable alignment across conditions, but this result should be interpreted descriptively rather than as independent evidence that anchoring is absent. These results suggest that assistance acted mainly as an editable temporal scaffold in this setting. However, they do not prove that all meaningful disagreement was preserved; rather, assistance improved efficiency and standardization without an observed large aggregate collapse in semantic-label entropy under the reported diagnostics.

Our broader contribution is an audit methodology: preserve disaggregated traces, separate efficiency from disagreement collapse, avoid treating consensus as ground truth, decompose boundary and semantic disagreement, and analyze whether model suggestions are accepted or actively revised. AI-assisted annotation should be evaluated not only by whether it makes labels faster or cleaner, but by how it changes the distribution of human judgments from which datasets are built.

\section*{Impact Statement}

This work studies AI-assisted data collection for high-ambiguity video annotation. Such workflows can reduce annotation effort, but they can also reshape the distribution of human judgments by anchoring annotators to model-generated suggestions. In pluralistic settings, this creates a risk that datasets become more standardized while losing disagreement that reflects meaningful differences in interpretation, context, or values.

We argue that AI-assisted annotation systems should preserve disaggregated annotator traces, document participant recruitment and task conditions, and report disagreement-related metrics rather than only aggregate labels or time savings. Such reports should also document how consensus references are constructed, whether evaluated annotations are included in those references, and how continuous annotations are discretized for metric computation. These practices can help researchers identify when AI assistance supports human judgment and when it suppresses it.

The video domains considered here may include sensitive, violent, or distressing content. Researchers and practitioners using similar workflows should provide content warnings, opt-out mechanisms, appropriate support, and fair compensation. The workflow should not be used to justify deployment of surveillance or anomaly-detection systems without broader stakeholder engagement and domain ethical review.

\bibliography{main}
\bibliographystyle{icml2026}

\newpage
\appendix
\onecolumn

\section{Fixed Pre-Annotation Engine and Model Validation}
\label{appendix:encoder_eval}

This appendix describes the Pre-Annotation engine used in the Assisted condition. The engine provides a fixed source of model-generated temporal suggestions for the annotation audit.

\subsection{Video-Text Encoding}

We employ a cross-modal retrieval architecture adapted for video downstream tasks \citep{radford_learning_2021}. Given a set of video-text pairs \((V, P)\), each video clip \(V_i \in \mathbb{R}^{T \times H \times W \times C}\) is treated as a sequence of \(T\) frames. Each frame is encoded to produce frame-level features \(z_i \in \mathbb{R}^{T \times D}\). A video-level embedding \(v_i \in \mathbb{R}^{D}\) is obtained by average pooling over the \(T\) frames \citep{rasheed2023fine}.

The corresponding text description \(p_i\) is processed by the text encoder to produce an embedding \(t_i \in \mathbb{R}^{D}\) in the shared feature space. The alignment between modalities is driven by cosine similarity \(\operatorname{sim}(v_i,t_i)\). We fine-tune the encoders using a contrastive objective. For a mini-batch of size \(B\), the video-to-text loss is:
\begin{equation}
\mathcal{L}_{v2t}
= -\frac{1}{B}\sum_{i=1}^{B}
\log
\frac{
\exp\!\big(\operatorname{sim}(v_i,t_i)/\tau\big)
}{
\sum_{j=1}^{B}\exp\!\big(\operatorname{sim}(v_i,t_j)/\tau\big)
}.
\end{equation}

The total loss averages the video-to-text and text-to-video directions:
\begin{equation}
\mathcal{L}_{\text{total}} = \frac{1}{2} \left( \mathcal{L}_{v2t} + \mathcal{L}_{t2v} \right).
\end{equation}

\begin{figure}[t]
    \centering
    \begin{adjustbox}{width=0.4\linewidth,center}
    \begin{tikzpicture}[
      node distance=0.9cm,
      every node/.style={font=\sffamily\small, align=center},
      box/.style={draw, rectangle, rounded corners, minimum width=2.2cm, minimum height=0.8cm, fill=white},
      dashedbox/.style={draw, rectangle, rounded corners, minimum width=2.2cm, minimum height=0.8cm, fill=gray!20, dashed},
      smallbox/.style={draw, rectangle, minimum width=2.2cm, minimum height=0.6cm},
      label/.style={font=\sffamily\footnotesize, align=center},
      arrow/.style={-Stealth, thick}
    ]
    \node[label] (image_seq) {\includegraphics[width=3cm]{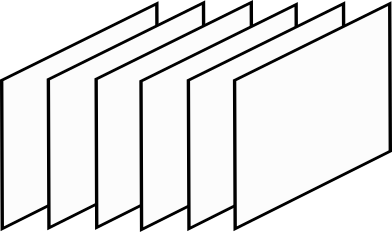}};
    \node[box, above=of image_seq] (image_enc) {Image Encoder\\trained};
    \node[label, above=of image_enc] (temp_enc) {\includegraphics[width=3.5cm]{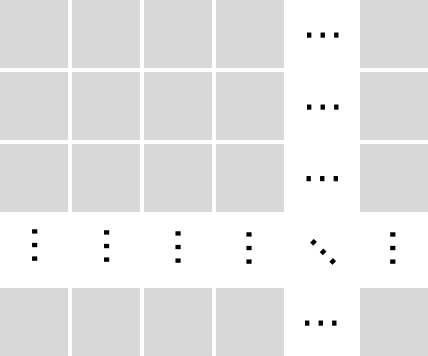}};
    \node[dashedbox, above=of temp_enc] (temp_pool) {Temporal Pooling};
    \node[dashedbox, above=of temp_pool, xshift=2.1cm, yshift=0.2cm] (max_sim) {Maximize Similarity};

    \node[label, right=of temp_enc, xshift=-0.5cm] (encoded_text) {\includegraphics[width=3.5cm]{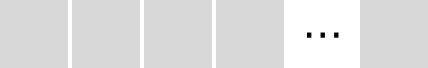}};
    \node[box, below=of encoded_text, yshift=-1.2cm] (text_enc) {Text Encoder\\trained};
    \node[smallbox, below=of text_enc, yshift=-0.4cm] (token) {Tokenized\\Prompt};

    \draw[arrow] (image_seq) -- node[right, pos=0.5] {Frames} (image_enc);
    \draw[arrow] (image_enc) -- node[right, pos=0.5] {$T \times D$} (temp_enc);
    \draw[arrow] (temp_enc) -- (temp_pool);
    \draw[arrow] (temp_pool) -- node[right, pos=0.5] {$1 \times D$} ++(0, 1) -| (max_sim);
    \draw[arrow] (token) -- (text_enc);
    \draw[arrow] (text_enc) -- node[right, pos=0.5] {$1 \times D$} (encoded_text);
    \draw[arrow] (encoded_text) -- ++(0, 3.93) -| (max_sim);
    \end{tikzpicture}
    \end{adjustbox}
    \caption{Cross-modal encoder used by the Pre-Annotation engine. Video frames are encoded, pooled into a video-level embedding, and aligned with text embeddings through contrastive training.}
    \label{diag:fine_tune}
\end{figure}

Figure~\ref{diag:fine_tune} illustrates the fixed cross-modal architecture used to produce the representations for Pre-Annotation. The diagram clarifies that temporal information enters through frame-level encoding followed by pooling, while supervision is provided through video-text contrastive alignment. This supports the interpretation of the engine as a semantic proposal generator rather than a frame-perfect temporal annotator.

\subsection{Hierarchical Content Modeling}

To generate temporal suggestions, we model the video feature space as a hierarchical tree. This structure supports multiple levels of granularity, from broad scene changes to specific actions. We construct the tree by recursively clustering feature arrays using spherical $k$-means \citep{JSSv050i10}. This variant maximizes cosine similarity between data points and centroids, matching the hyperspherical structure of normalized CLIP embeddings \citep{liang2022mind}. The objective is:
\begin{equation}
\max_{r_{ij},\,\boldsymbol{\mu}_j}
\sum_{i=1}^{p}
\sum_{j=1}^{k}
r_{ij}\,
\frac{
\boldsymbol{\mu}_j \cdot \mathbf{x}_i
}{
\lVert \boldsymbol{\mu}_j \rVert
\;\lVert \mathbf{x}_i \rVert
},
\end{equation}
subject to \(r_{ij}\in\{0,1\}\) and \(\sum_{j=1}^{k} r_{ij}=1\) for all \(i\). Here, \(\boldsymbol{\mu}_j\) is the centroid of cluster \(j\), \(\mathbf{x}_i\) is the \(i\)-th data point, and \(r_{ij}=1\) indicates assignment of \(\mathbf{x}_i\) to cluster \(j\).

The process begins with high-level clusters and recursively refines them into fine-grained nodes. We use a maximum tree depth \(T_{\max}=3\) and a semantic cohesion threshold \(\tau=0.85\), halting subdivision when within-cluster cosine similarity exceeds the threshold. Leaf nodes become Pre-Annotation segments.

\begin{figure}[t]
    \centering
    \includegraphics[width=\textwidth]{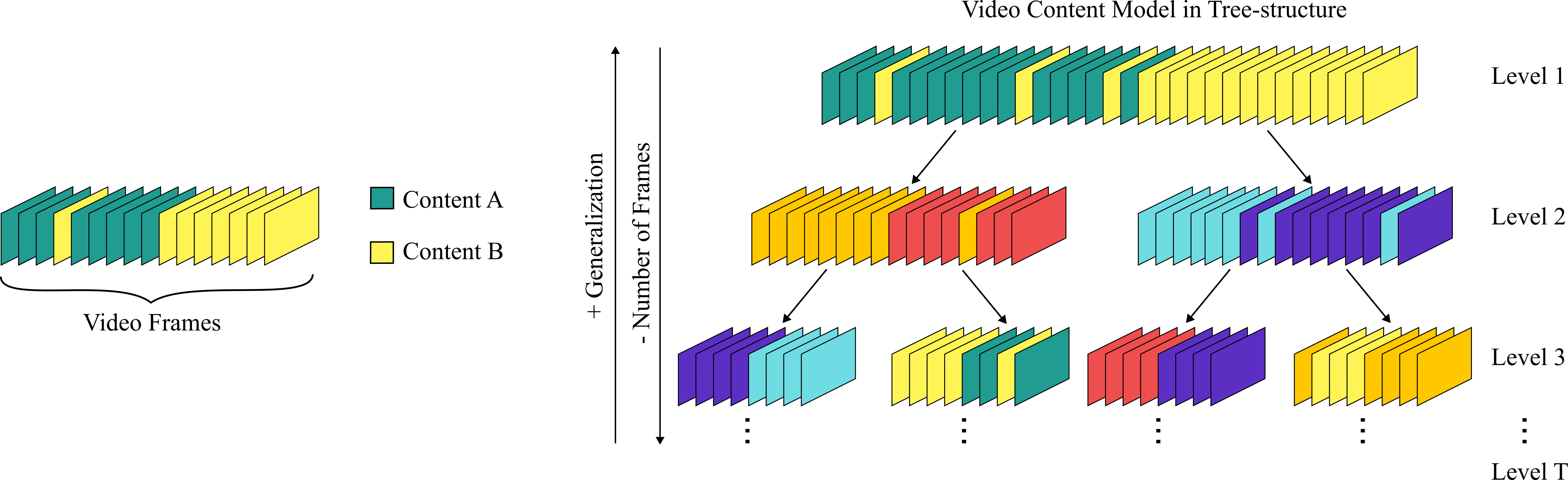}
    \caption{Hierarchical clustering of video content. Frames are recursively clustered into a tree structure using spherical $k$-means. Leaf nodes are converted into Pre-Annotation segments.}
    \label{fig:video_content_structure}
\end{figure}

\subsection{Training Datasets and Setup}

We trained the encoder using a composite of four datasets to expose the model to diverse visual environments, action durations, and language descriptions. UCF-Crime provides 1,900 untrimmed real-world surveillance videos totaling 128 hours \citep{Sultani_2018_CVPR}. UCA augments UCF-Crime with 23,542 linguistic annotations across 110.7 hours of footage \citep{wu2023vadclip}. XD-Violence contributes 4,754 untrimmed videos totaling 217 hours \citep{Wu2020not}. ActivityNet Captions adds 849 hours of footage across 20,000 videos \citep{krishna2017}.

The cross-modal encoder was fine-tuned on an NVIDIA A100 GPU with 40 GB VRAM. We used AdamW \citep{loshchilov2017decoupled} with a decoupled weight decay of \(1 \times 10^{-3}\). The initial learning rate was \(8 \times 10^{-6}\). We used cosine learning-rate decay \citep{loshchilov2017sgdrstochasticgradientdescent}:
\begin{equation}
\label{equation:lr}
\mathrm{lr}(t)
=
\frac{1}{2}\,\mathrm{lr}_{\mathrm{base}}
\left(
1
+
\cos\!\left(
\frac{t - t_{\mathrm{warm\text{-}up}}}
     {T_{\mathrm{total}} - t_{\mathrm{warm\text{-}up}}}
\,\pi
\right)
\right),
\end{equation}
where \(T_{\text{total}}\) is the total number of training steps and \(t_{\text{warm-up}}\) is the number of warm-up steps.

\subsection{Evaluation Protocol and Results}

We evaluated the encoder using video-text matching. For each text query, the model ranked the matching video clip against a candidate pool of approximately 12--13 clips, consisting of the positive clip and randomly sampled negative clips. This candidate-pool size is consistent with the random baseline in Table~\ref{table:clip_fine-tune}. The model predicts the most probable video-text pair by comparing embeddings with cosine similarity and normalizing scores with a softmax. During evaluation, we introduced randomness by sampling negative clips both from within the same source video and from different videos, making the retrieval task include visually similar distractors as well as cross-video distractors. To reduce leakage, all clips derived from the same source video were assigned to the same train, validation, or test split.

We report Accuracy at Rank \(\mathbb{K}\) (Acc@\(\mathbb{K}\)). Acc@\(\mathbb{K}\) is the percentage of test samples for which the correct result appears within the top-\(\mathbb{K}\) retrieved clips. Table~\ref{table:clip_fine-tune} reports Acc@1 and Acc@5 for several configurations.

\begin{table}[h]
\caption{Performance comparison of CLIP fine-tuning configurations.}
\label{table:clip_fine-tune}
\vskip 0.15in
\begin{center}
\begin{small}
\begin{sc}
\begin{tabular}{lcccc}
\toprule
Backbone & Batch Size & Sampled Frames & Acc@1 (\%) & Acc@5 (\%) \\
\midrule
Random & -- & -- & 8.17 & 39.89 \\
\midrule
\multirow{2}{*}{ViT-B/16}
& 10 & 16 & \textbf{89.68} & \textbf{99.64} \\
& 30 & 30 & 73.59 & 95.30 \\
\midrule
\multirow{4}{*}{ViT-B/32}
& 5 & 30 & 73.71 & 95.20 \\
& 30 & 5 & 73.71 & 95.20 \\
& 20 & 32 & 81.76 & 97.75 \\
& 100 & 40 & 59.59 & 86.24 \\
\bottomrule
\end{tabular}
\end{sc}
\end{small}
\end{center}
\vskip -0.1in
\end{table}

The selected ViT-B/16 configuration with batch size 10 and 16 sampled frames achieved the highest Acc@1 score, 89.68\%, and Acc@5 score, 99.64\%. This configuration was used to generate the Pre-Annotations in the Assisted condition.

\subsection{Attention Visualization}
\label{appendix:attention_shift}

We visualized self-attention maps from the Vision Transformer to inspect whether fine-tuning shifted attention toward relevant regions in representative frames \citep{NIPS2017_3f5ee243,radford_learning_2021}. We extracted multi-head attention weights, computed cumulative joint attention across layers, and isolated attention associated with the CLS token. Attention maps were averaged across heads and upscaled to the input frame resolution \citep{Ma2022Visualizing,Li2023How}.

\begin{figure}[t]
\centering
\setlength{\fboxsep}{0pt}
\newlength{\mergedwd}
\setlength{\mergedwd}{0.85\linewidth}
\newcommand{\labelxshift}{-0.3cm}

\begin{tikzpicture}
  \node[inner sep=0] (img) {\includegraphics[width=\mergedwd]{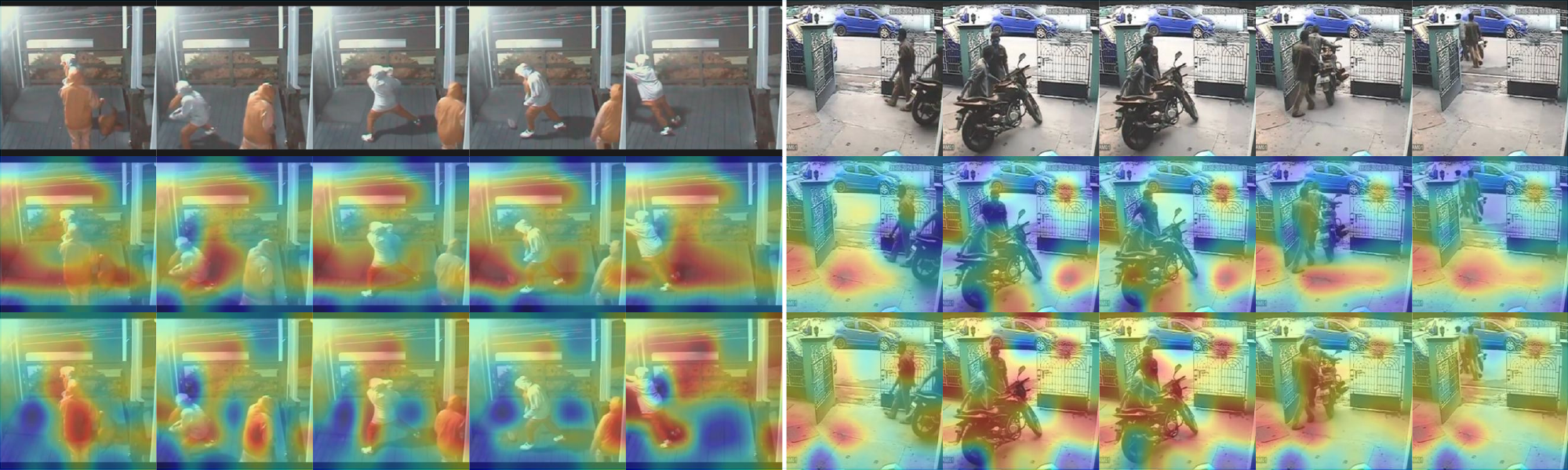}};
  \node[rotate=90, anchor=center] at ($(img.north west)!1/6!(img.south west) + (\labelxshift,0)$) {\small Original};
  \node[rotate=90, anchor=center] at ($(img.north west)!1/2!(img.south west)  + (\labelxshift,0)$) {\small ViT-B/32};
  \node[rotate=90, anchor=center] at ($(img.north west)!5/6!(img.south west) + (\labelxshift,0)$) {\small Trained};
\end{tikzpicture}

\vspace{1mm}

\noindent\parbox{\mergedwd}{%
  \centering
  \begin{tabular}{@{} p{\dimexpr0.48\mergedwd\relax} @{\hspace{0.02\mergedwd}} p{\dimexpr0.48\mergedwd\relax} @{} }
    \centering\small\textbf{Vandalism034}: ``A group of people vandalizes a black trash can by repeatedly kicking, pushing, and hitting it.'' &
    \centering\small\textbf{Stealing004}: ``Two people coordinate to push a motorcycle out of a yard and onto a road.'' \\
  \end{tabular}
}

\vspace{4mm}

\begin{tikzpicture}
  \node[inner sep=0] (img2) {\includegraphics[width=\mergedwd]{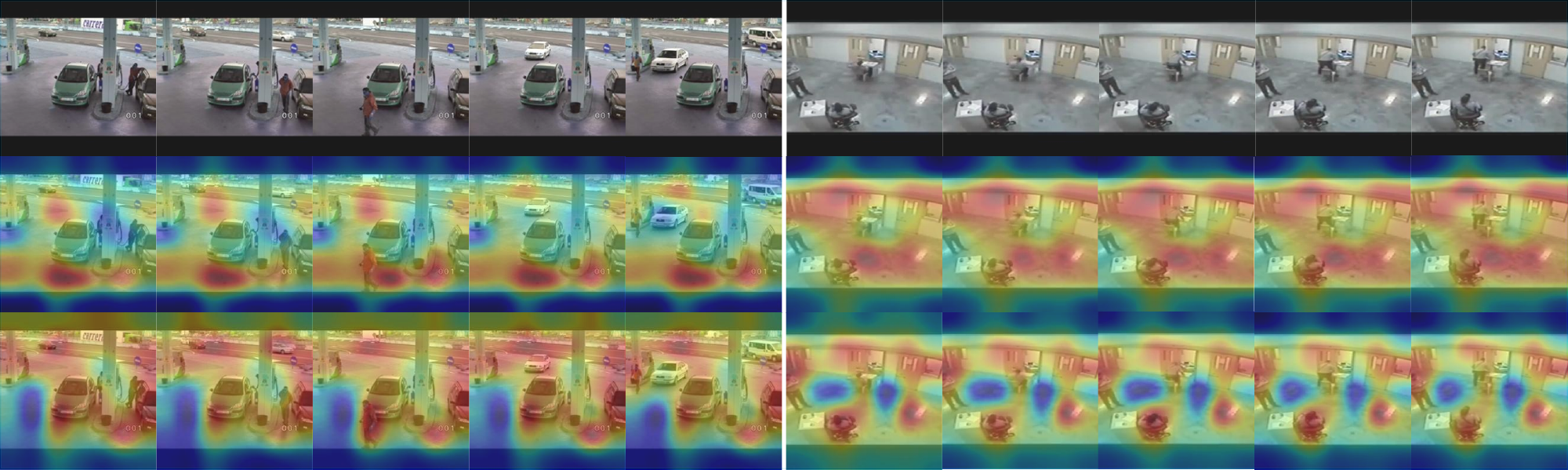}};
  \node[rotate=90, anchor=center] at ($(img2.north west)!1/6!(img2.south west) + (\labelxshift,0)$) {\small Original};
  \node[rotate=90, anchor=center] at ($(img2.north west)!1/2!(img2.south west)  + (\labelxshift,0)$) {\small ViT-B/32};
  \node[rotate=90, anchor=center] at ($(img2.north west)!5/6!(img2.south west) + (\labelxshift,0)$) {\small Trained};
\end{tikzpicture}

\vspace{1mm}

\noindent\parbox{\mergedwd}{%
  \centering
  \begin{tabular}{@{} p{\dimexpr0.48\mergedwd\relax} @{\hspace{0.02\mergedwd}} p{\dimexpr0.48\mergedwd\relax} @{} }
    \centering\small\textbf{Normal\_Videos\_352}: ``People refuel cars and leave a gas station.'' &
    \centering\small\textbf{Abuse005}: ``A guard moves around a room and interacts with another person.'' \\
  \end{tabular}
}

\caption{Attention map visualization on representative UCF-Crime samples. Each merged image contains three stacked rows: original frame, baseline ViT-B/32 attention, and fine-tuned attention.}
\label{figure:ViT_attention_maps}
\end{figure}

Figure~\ref{figure:ViT_attention_maps} provides a qualitative sanity check on the fine-tuned encoder. The trained attention maps appear to concentrate more on task-relevant actors, objects, or interaction regions than the baseline maps in these examples. We use this visualization only as qualitative support; the quantitative validation of the encoder is provided by the retrieval results in Table~\ref{table:clip_fine-tune}.

\section{Annotation Interface}
\label{appendix:interface}

\begin{figure}[h]
    \centering
    \begin{subfigure}[b]{0.95\linewidth}
        \centering
        \includegraphics[width=\linewidth]{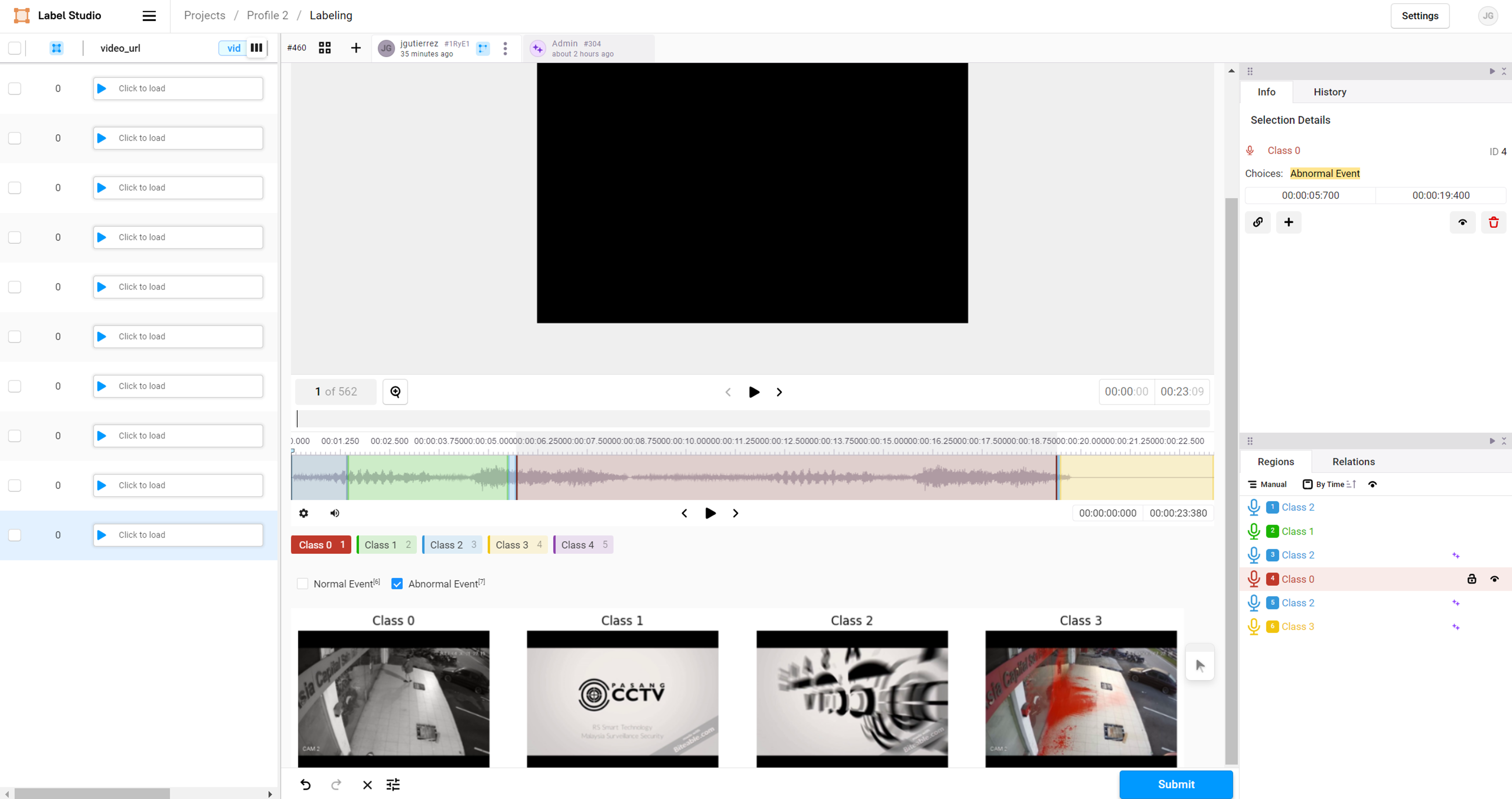}
        \caption{Main annotation workspace with video player and class controls.}
        \label{fig:ls_main}
    \end{subfigure}

    \vspace{0.2cm}

    \begin{subfigure}[b]{0.95\linewidth}
        \centering
        \includegraphics[width=\linewidth]{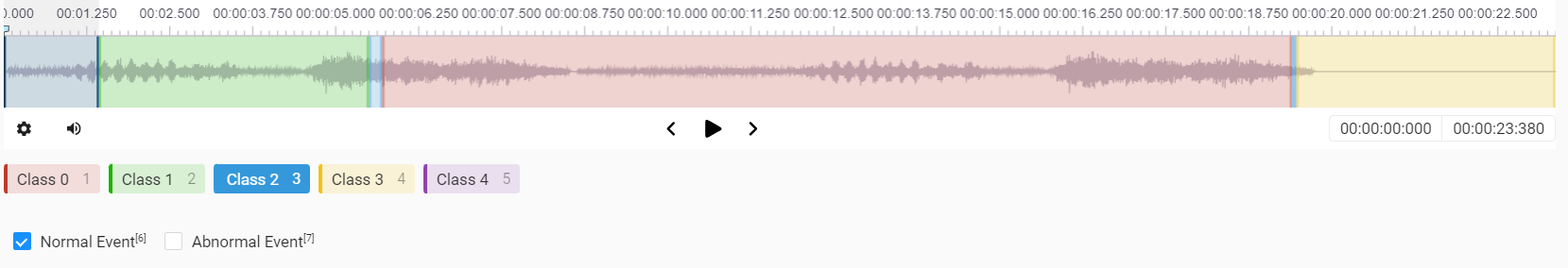}
        \caption{Temporal timeline with suggested segments and binary event controls.}
        \label{fig:ls_timeline}
    \end{subfigure}

    \caption{Customized Label Studio interface. In the Assisted condition, Pre-Annotations appear on the temporal bar as suggestions that annotators can copy, edit, split, merge, delete, or relabel.}
    \label{fig:interface_split}
\end{figure}

Figure~\ref{fig:interface_split} also clarifies the behavioral intervention studied in the main paper. In the Unassisted condition, annotators construct the temporal segmentation from scratch using the same workspace. In the Assisted condition, the temporal bar is pre-populated with editable suggestions, shifting the task from initial construction toward review, correction, splitting, merging, and relabeling.

\section{CrowdWorkSheets Documentation}
\label{appendix:CrowdWorkSheets}

We include a CrowdWorkSheets-style description of the annotation process \citep{diaz2022crowdworksheets}. The purpose is to document task subjectivity, annotator recruitment, platform choices, and known risks.

\subsection{Task Formulation}

\paragraph{At a high level, what are the subjective aspects of your task?}
The task is subjective in two ways. First, annotators decide how to group frames into temporally coherent events. The boundary between the end of one event and the beginning of the next can be gradual. Second, annotators classify each segment as normal or abnormal. The guide defined abnormal events as criminal, accidental, or catastrophic incidents, but applying this definition still requires contextual interpretation.

\paragraph{What assumptions do you make about annotators?}
We assume that annotators follow the instructions, watch each video before annotating, and use their own judgment when boundaries or event categories are ambiguous. We also assume that disagreements are expected and informative in this task.

\paragraph{How did you choose the wording of the task instructions?}
The instructions were designed to give annotators a concrete operational definition while leaving room for judgment. The abnormal-event definition was intentionally broad enough to cover the relevant video categories without forcing frame-perfect agreement. The guide was reviewed by multiple researchers and updated after early annotation batches.

\paragraph{What risks did the task pose for annotators?}
The task involved videos that could include violent, explicit, catastrophic, or otherwise distressing content. Annotators were informed of the content type before participation. Videos were filtered before the study to remove material judged too intense for the annotation setting.

\paragraph{What precise instructions were provided to annotators?}
Annotators were asked to annotate open-world videos. Each annotator labeled 10 videos: 5 without Pre-Annotations and 5 with model-generated Pre-Annotations. For each video, annotators grouped frames with similar content by marking temporal segments and assigning class labels. They also marked each segment as normal or abnormal. Annotators were instructed to watch the full video before beginning annotation and to use the guide when applying the normal/abnormal distinction.

\subsection{Selecting Annotators}

\paragraph{Were certain perspectives privileged?}
The study used a controlled university-affiliated cohort. The aim was to measure how the annotation workflow changes behavior under a consistent study protocol.

\paragraph{Were sociodemographic characteristics used to select annotators?}
Annotators were volunteers affiliated with a university institution. They had at least university-level academic background and no prior exposure to the specific test clips. The cohort was balanced in gender and included ages ranging from 18 to 50.

\paragraph{Could annotator characteristics have affected the annotations?}
Yes. Judgments about normality, abnormality, event boundaries, and contextual relevance can vary across communities and backgrounds. This is why the main paper treats the study as a controlled annotation audit and does not interpret the Consensus Reference as objective ground truth.

\paragraph{Are affected communities represented in the annotator pool?}
The annotator pool is not designed to represent all communities that may be affected by video analysis systems. The study evaluates the annotation workflow and preserves disaggregated traces for analysis. Broader deployment would require stakeholder-specific recruitment and analysis.

\subsection{Platform and Infrastructure Choices}

\paragraph{What annotation platform did you use?}
We used Label Studio \citep{Labelstudio}.

\paragraph{Why was this platform chosen?}
Label Studio is open source, configurable, and supports temporal video annotation. It also records timing information and allows integration with model-generated Pre-Annotations.

\paragraph{Did the platform meet the requirements of the task?}
Yes. The customized interface supported video playback, temporal segmentation, segment labeling, binary event flags, and Pre-Annotation review.

\paragraph{What communication channels were available?}
Annotators could ask clarification questions directly or through email. Questions about tool usage were answered. Questions about ambiguous content were answered by referring annotators back to the written guide and the principle that reasonable disagreement was expected.

\paragraph{How were annotators compensated?}
Annotators participated voluntarily and were not financially compensated.

\subsection{Dataset Analysis and Evaluation}

\paragraph{How do you define annotation quality in this context?}
Quality is not defined as agreement with a single ground truth. We evaluate the annotation process through multiple diagnostics: annotation time, alignment with an aggregate Consensus Reference, pairwise inter-annotator consistency, and latent-space standardization.

\paragraph{Did you analyze disagreement patterns?}
Yes. Section~\ref{sec:evaluation} analyzes disagreement through pairwise inter-annotator consistency and compares Assisted and Unassisted conditions. It also measures latent-space standardization to distinguish boundary regularization from simple agreement increase.

\paragraph{How do individual annotator responses relate to released labels?}
The released artifact preserves annotator-level traces rather than only a final aggregated label. This allows future work to study disagreement, revision behavior, and the effect of model suggestions on the annotation process.

\paragraph{Ethics and participant protection.}
Participants were informed before the study that the videos could contain violent, abnormal, catastrophic, or otherwise distressing content. Participation was voluntary, and participants could stop the task at any time. The study used a controlled university-affiliated cohort and filtered videos before annotation to remove material judged too intense for the study setting. Because unpaid annotation of distressing content raises ethical concerns, we treat the volunteer design as a limitation of the pilot and recommend fair compensation, explicit opt-out mechanisms, and debriefing resources for larger-scale studies.

\end{document}